\pdfoutput=1

\documentclass[11pt]{article}

\usepackage[final]{acl}

\usepackage{times}
\usepackage{latexsym}
\usepackage{amsmath}
\usepackage{tcolorbox}
\usepackage{graphicx}
\usepackage{booktabs}
\usepackage[T1]{fontenc}

\usepackage[utf8]{inputenc}

\usepackage{microtype}

\usepackage{inconsolata}
\usepackage{enumitem}
\usepackage{xcolor}

\usepackage{graphicx}
\usepackage{algorithm}
\usepackage{algpseudocode}

\title{Decompose-ToM: Enhancing Theory of Mind Reasoning in Large Language Models through Simulation and Task Decomposition}

\author{Sneheel Sarangi \\
  NYU Abu Dhabi \\
  \texttt{sneheelsarangi@nyu.edu} \\\And
  Maha Elgarf \\
  NYU Abu Dhabi\\
  \texttt{maha.elgarf@nyu.edu}
  \\\And
  Hanan Salam \\
  NYU Abu Dhabi \\
  \texttt{hanan.salam@nyu.edu} \\}

\begin{document}
\maketitle
\begin{abstract}
Theory of Mind (ToM) is the ability to understand and reflect on the mental states of others. Although this capability is crucial for human interaction, testing on Large Language Models (LLMs) reveals that they possess only a rudimentary understanding of it. Although the most capable closed-source LLMs have come close to human performance on some ToM tasks, they still perform poorly on complex variations of the task that involve more structured reasoning. In this work, we utilize the concept of "pretend-play", or ``Simulation Theory'' from cognitive psychology to propose ``Decompose-ToM'': an LLM-based inference algorithm that improves model performance on complex ToM tasks. We recursively simulate user perspectives and decompose the ToM task into a simpler set of tasks: subject identification, question-reframing, world model updation, and knowledge availability. We test the algorithm on higher-order ToM tasks and a task testing for ToM capabilities in a conversational setting, demonstrating that our approach shows significant improvement across models compared to baseline methods while requiring minimal prompt tuning across tasks and no additional model training. Our code is \href{https://github.com/Xarangi/Decompose-ToM}{publicly available}.
\end{abstract}

\section{Introduction}

Social reasoning, the ability to understand and navigate complex interpersonal dynamics and social contexts, is crucial for large language models (LLMs) as it both enables better interactions with users \cite{salam2022learning, salam2023automatic}, and paves the way for more powerful LLM-based systems that may act in socially appropriate and helpful ways within the same environment as humans \cite{li2023survey,jiang2024towards}. A key factor in the human ability to socially reason is the presence of the theory of mind (ToM) \cite{BaronCohen1995, Christensen2012}. ToM is the ability to attribute and infer the mental states of others, a capability children start developing at a young age \cite{Premack1978}. However, the performance of LLMs on this task is contentious. In the ToMi \cite{le-etal-2019-revisiting} dataset, some recent methods have improved the capabilities of frontier models to near-human levels \cite{sclar-etal-2023-minding,wilf-etal-2024-think}. At the same time, the research community has designed newer datasets with variations to the ToM task that continue to be a challenge for LLMs.

For example, higher-order ToM capabilities involve the ability to model the beliefs of others about the beliefs of others and so on \cite{Kinderman1998}. Studies have highlighted the pivotal role of higher-order Theory of Mind (ToM) not only in managing intricate communication scenarios such as discussions involving multiple parties \cite{HitomSocialCompetenceChildren}, but also in fostering empathetic interactions and providing emotional support \cite{OverlappingRelationshipEmotionPerception&TOM}. However, LLM models continue to perform poorly in tasks that aim to measure this capability \cite{wu-etal-2023-hi}. Likewise, LLM models struggle to generalize even lower-order Theory of Mind (ToM) reasoning demonstrated in datasets like ToMi to more realistic dialogue-based scenarios, as observed by \citet{kim-etal-2023-fantom}. This raises the question: Can LLMs or LLM-based approaches effectively address these challenges while maintaining task generalization?

Insights from developmental psychology offer potential solutions. The development of ToM in children has been closely linked to the act of ``pretend play'' \cite{Kavanaugh2006-dk,Lillard1993-pw,Qu2015}. Emerging around 18 months of age---significantly earlier than the age at which children reliably solve Theory of Mind (ToM) tasks like those in ToMi---this process involves role-taking and the attribution of mental states, both of which are essential for developing ToM capabilities \cite{Leslie1987}. Similarly, another precursor to ToM capabilities in children is the ability to solve ``knowledge-access''. Children as young as 3 years of age begin to understand the connection between perception and knowledge, recognizing that visual access to information leads to knowing \cite{Wellman2004}.

Parallel insights can also be drawn from computational approaches to reasoning, particularly the recursive reasoning frameworks employed in Rational Speech Act (RSA) models. RSA models conceptualize communication as a recursive process in which speakers anticipate listeners’ interpretations and listeners infer speakers’ intentions \cite{frank2012predicting}. This recursive reasoning mirrors the structure of higher-order ToM tasks, where agents must model others’ beliefs about beliefs. Furthermore, RSA models address challenges in scalar implicature by recursively balancing informativeness and cost—an approach analogous to how ToM reasoning involves balancing observed actions and inferred mental states. By grounding ToM reasoning in recursive simulation, we can enhance the ability of LLMs to reason about complex mental states, paving the way for better performance on higher-order tasks and naturalistic interactions.

In this paper, we propose an LLM-powered inference time algorithm - Decompose-ToM - that takes inspiration from the above concepts to decompose the ToM reasoning problem into two simpler problems - recursively simulating an agent's perspectives, and using granular knowledge-access problems to simulate these perspectives. Given a question and a story, at each step, the algorithm simplifies the story to a set of statements known only to the agent being simulated, until the question no longer requires ToM-based inference and simply asks for a factual answer. We demonstrate that our method significantly outperforms baselines while requiring no further training and maintaining generality across datasets.

\section{Related Work}

\begin{figure*}[t]
    \centering
    \includegraphics[width=\textwidth]{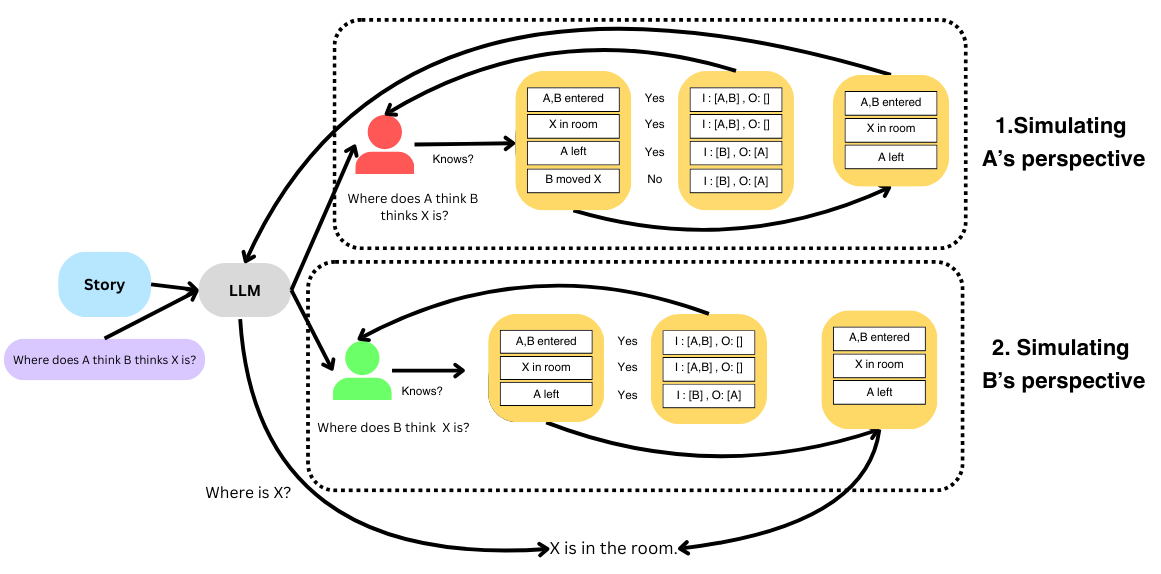}
    \caption{Workflow of Decompose-ToM. The perspectives of A and B are simulated sequentially. Then, a factual question is answered based on a simplified story.}
    \label{fig:1}
\end{figure*}

\textbf{Towards a Machine Theory of Mind}

The pursuit of a machine Theory of Mind (ToM) has been a significant focus within the research community for many years. Early investigations utilizing neural language models revealed that these models often faced challenges in accurately inferring mental states \cite{nematzadeh-etal-2018-evaluating}. As large language models (LLMs) have advanced, researchers have increasingly turned their attention to assessing these models' ToM capabilities. For example, \citet{sap-etal-2022-neural} employed versions of the Sally-Anne false-belief test \cite{Baron-Cohen-Autistic-Child-Have-ToM?} using the ToMi \cite{le-etal-2019-revisiting} dataset to illustrate the limited performance of LLMs like GPT-3 on tasks that require belief inference.

In contrast, more recent and sophisticated models such as GPT-4 \cite{openai2023gpt4} have demonstrated performance that approaches or even surpasses human-level accuracy on benchmarks like ToMi \cite{kosinski2023theory} and BigToM \cite{gandhi2023understanding}. Despite these promising developments, concerns persist regarding the robustness of LLMs' ToM capabilities. Notably, the effectiveness of these models tends to diminish significantly when subjected to adversarial perturbations \cite{ullman2023largelanguagemodelsfail,shapira2023cleverhansneuraltheory}, raising questions about the consistency and reliability of their mental state inferences.

\textbf{Strategies for ToM Performance in LLMs}. Traditional methods that have worked to improve LLM performance on reasoning tasks such as Chain of Thought prompting \cite{wei2022chain}, have seen limited success in improving LLMs' ToM performance \cite{moghaddam2023boosting}. There have been recent works such as Symbolic-ToM \cite{sclar-etal-2023-minding}, and SimToM \cite{wilf-etal-2024-think} attempting to improve the performance of LLMs on false-belief ToM tasks. Symbolic-ToM utilizes LLMs to form a belief-state graph representation, before attempting to answer a question. This method provides an elegant solution for even higher-order ToM tasks, but it increases the algorithm's complexity exponentially. Additionally, the method of adapting it to more naturalistic settings remains unclear. SimToM proposes a simple 2-step method involving perspective-taking, taking inspiration from Simulation Theory \cite{Shanton2010}. While drawing from similar inspiration in its design philosophy, SimToM doesn't consider more complex ToM scenarios such as higher-order reasoning.

\textbf{Evaluation of ToM in LLMs }. Given the saturation of benchmarks such as ToMi due to the improved performance of newer models, recent research has moved towards testing more complex variations of the classical false-belief tasks, such as tests for performance on higher-order ToM tasks \cite{wu-etal-2023-hi}, and in naturalistic dialogue settings \cite{kim-etal-2023-fantom}. There have also been recent datasets to test LLM performance on a broad array of ToM tasks such as treating LLMs as agents embedded in environments, and other psychology-based ToM tests such as faux-pas tests \cite{ma-etal-2023-towards-holistic,xu-etal-2024-opentom}.

\section{Methodology}

Drawing inspiration from the concepts of pretend-play \cite{Kavanaugh2006-dk,Lillard1993-pw,Qu2015} and knowledge-access \cite{Wellman2004}, in a similar vein to SimToM \cite{wilf-etal-2024-think}, our approach ``Decompose-ToM'' decomposes the ToM task into a recursive perspective simulation task, and a granular statement awareness task. Concretely, this means that for each statement describing a theory of mind testing scenario, we prompt the LLM to output whether an interested agent is aware of it. Additionally, to assist the LLM in this task, we keep track of a symbolic world state representation that is both generated and updated by the LLM. We recursively chain agent simulations, simplifying the question to be answered at each stage until we get a question that is answered by a fact, and not an agent's belief. This process is demonstrated in Appendix \ref{app:algos} as Algorithm \ref{alg:SimulationProcess}, Algorithm \ref{alg:StoryUpdate}, and Figure \ref{fig:1}. Our approach can thus be divided into three components: (1) Initialization, (2) Simulation Processing \& World-state Updating, and (3) Question-Answering\footnote{The reader is referred to Appendix \ref{app:prompts} for a detailed description of the prompts used in this work.}.

\paragraph{Initialization.}

This step extracts the character whose perspective will be simulated and then simplifies the question to frame it as if it were asked to the character. We accomplish both these tasks by few-shot prompting the LLM with hand-crafted demonstrations. For example, if the question states: ``Where does A think B thinks the object O is?'' We detect the person to be simulated as A, and the simplified question would thus be: ``Where does B think the object O is?''. At this step we also set up the text-based world state by prompting the LLM to detect the locations mentioned in a story, and output these arranged into a simple world state representation such as: ``Living Room:[], Bedroom:[]''. 

\paragraph{Simulation Processing \& Updating World State.}

This step creates a new story consisting of only statements known by the identified character from the overall story. For each statement, we prompt the LLM to answer whether the given character knows the statement by inferring from prior statements in the story and the given world-state representation. After the LLM makes a decision, we update the world state with another LLM call if it needs to be updated after the given statement. If the LLM decides that the given statement is known by the agent, we add the statement to the sub-story. After iterating through all the statements in the story, we obtain a newly created story that represents the story from the perspective of the identified character.

\paragraph{Question-Answering.}

The initialization and processing steps above are followed recursively until the character-identification step recognizes the question to have decomposed into a factual question. Thus, it no longer requires the theory-of-mind process to be answered and can be answered directly. Then, we prompt the LLM to answer the question using the simplified story in a chain-of-thought manner, asking it to form a reasoning and then having it select from the multiple-choice options.

\begin{table*}[ht]
\footnotesize
\centering
\begin{tabular}{|l|c|c|c|c|c|c|}
\hline
\textbf{Dataset} & \multicolumn{4}{c|}{\textbf{Hi-ToM}} & \multicolumn{2}{c|}{\textbf{FANToM}} \\
\hline
\textbf{Model} & \textbf{Order 1} & \textbf{Order 2} & \textbf{Order 3} & \textbf{Order 4} & \textbf{Short} & \textbf{Long} \\
\hline
\multicolumn{7}{|c|}{\textbf{MC Prompt Baseline}} \\
\hline
\textbf{Llama3 8b} & 67.5 & \textbf{56.67} & 35.83 & 38.33 & 32.4 & 27.5 \\
\textbf{Llama3 70b} & 58.33 & 45 & 40.83 & 37.5 & 61.9 & 50.6 \\
\textbf{Gemini Flash} & 60 & 49.17 & 30 & 26.67 & 44.9 & 41.8 \\
\textbf{GPT 4o} & 60 & 40.83 & 22.5 & 21.67 & 55 & 44 \\
\hline
\multicolumn{7}{|c|}{\textbf{CoT}} \\
\hline
\textbf{Llama3 8b} & 58.33 & 50 & 36.67 & 37.50 & 57.1 & 45.6 \\
\textbf{Llama3 70b} & 68.33 & 54.17 & 44.17 & 39.17 & 68.4 & 61.4 \\
\textbf{Gemini Flash} & 55.83 & \textbf{52.5} & 45.83 & 34.17 & 39.2 & 36.1 \\
\textbf{GPT 4o} & 74.17 & 53.33 & 45 & 42.5 & 74 & 64.7 \\
\hline
\multicolumn{7}{|c|}{\textbf{SimToM}} \\
\hline
\textbf{Llama3 8b} & 57.5 & 46.67 & 39.17 & 31.67 & 60.09 & 60.33 \\
\textbf{Llama3 70b} & 67.5 & 52.5 & 55 & 43.33 & \textbf{89.25} & 83.37 \\
\textbf{Gemini Flash} & \textbf{68.33} & 51.67 & 45 & 46.67 & 70.91 & 66.87 \\
\textbf{GPT 4o} & 75 & 55 & 48.33 & 43.33 & \textbf{90.43} & 84.7 \\
\hline
\multicolumn{7}{|c|}{\textbf{Decompose-ToM (Our Method)}} \\
\hline
\textbf{Llama3 8b} & 
\textbf{69.2} \scriptsize{(\textcolor{green}{+1.7})} & 
50.8 \scriptsize{(\textcolor{red}{-5.8})} & 
\textbf{51.7} \scriptsize{(\textcolor{green}{+12.5})} & 
\textbf{55.0} \scriptsize{(\textcolor{green}{+16.7})} & 
\textbf{62.5} \scriptsize{(\textcolor{green}{+2.4})} & 
\textbf{62.0} \scriptsize{(\textcolor{green}{+1.7})} \\
           
\textbf{Llama3 70b} & 
\textbf{74.2} \scriptsize{(\textcolor{green}{+5.8})} & 
\textbf{76.7} \scriptsize{(\textcolor{green}{+22.5})} & 
\textbf{84.2} \scriptsize{(\textcolor{green}{+29.2})} & 
\textbf{87.5} \scriptsize{(\textcolor{green}{+44.2})} & 
86.8 \scriptsize{(\textcolor{red}{-2.4})} & 
\textbf{86.1} \scriptsize{(\textcolor{green}{+2.7})} \\
           
\textbf{Gemini Flash} & 
65.6 \scriptsize{(\textcolor{red}{-2.8})} & 
\textbf{51.7} \scriptsize{(\textcolor{red}{-0.8})} & 
47.8 \scriptsize{(\textcolor{green}{+2.0})} & 
\textbf{51.3} \scriptsize{(\textcolor{green}{+4.6})} & 
\textbf{75.63} \scriptsize{(\textcolor{green}{+4.7})} & 
\textbf{75.43} \scriptsize{(\textcolor{green}{+8.6})} \\
           
\textbf{GPT 4o} & 
\textbf{76.7} \scriptsize{(\textcolor{green}{+1.7})} & 
\textbf{86.7} \scriptsize{(\textcolor{green}{+31.7})} & 
\textbf{87.5} \scriptsize{(\textcolor{green}{+39.2})} & 
\textbf{83.3} \scriptsize{(\textcolor{green}{+40.0})} & 
88.4 \scriptsize{(\textcolor{red}{-2.0})} & 
\textbf{86.2} \scriptsize{(\textcolor{green}{+1.5})} \\
\hline
\end{tabular}
\caption{Performance of Various Models on Hi-ToM and FANToM in terms of accuracy (\%), with gains over the best performing method amongst MC Prompt Baseline, CoT, and SimToM.}
\end{table*}

\section{Experiments and Results}

\subsection{Datasets}
\paragraph{Hi-ToM \cite{wu-etal-2023-hi}.}
Hi-ToM is designed to evaluate the higher-order theory of mind abilities of language models, going up to the fourth-order theory of mind reasoning tasks. It is based on the Sally-Anne test \cite{Baron-Cohen-Autistic-Child-Have-ToM?} with characters entering, leaving, and moving items in rooms. The provided task is a multiple-choice question-answering task with 15 choices per question. The dataset consists of 600 questions spanning two categories: ``Tell'' and ``no-Tell''. The ''Tell'' category adds an element of deception, by asking LLMs to reason whether an agent is lying by inferring when they left the room. We present our results averaged over both categories.

\paragraph{FANToM \cite{kim-etal-2023-fantom}.}

FANToM is a dataset consisting of dialogue-based interactions between characters. Characters may leave or enter the conversation at any time, and questions are based on correctly navigating this information incompleteness. This provides a naturalistic setting to test our model on. While the dataset provides a large suite of tasks, we use the binary-choice belief task for our evaluations due to its structural similarity to the Hi-ToM dataset. The dataset consists of 1,540 questions, containing first and second-order ToM questions. We show averaged results over these categories due to generally similar accuracy gains for models across evaluation methods. Additionally, the dataset consists of stories in short and long-length variants, which we show results for separately to evaluate our method's performance over longer context lengths.

We provide example stories from both datasets, alongside details of any pre-processing conducted, in Appendix \ref{app:datasets}.

\subsection{Evaluation}

We experiment with four base LLMs, and evaluate each of them via: \newline
- \textbf{Zero-shot prompting (Baseline)}: Model directly returns the multiple choice option of the answer\newline
- \textbf{Zero-shot chain of thought prompting}: Model "thinks step-by-step" before outputting the answer\newline
- \textbf{SimToM prompting \cite{wilf-etal-2024-think}}: Two-step prompting, first involving perspective-taking, then finding the answer\footnote{For details on SimToM refer to Appendix \ref{simtom}}\newline
- \textbf{Our approach (Decompose-ToM)}: Decomposes tasks into recursive simulations and knowledge access subtasks. 

Each method is applied consistently to compare performance across models and datasets.
We evaluate the open-source models Llama-3-8B and 70B \cite{dubey2024llama3herdmodels}, alongside the closed source models Gemini-1.5-Flash \cite{geminiteam2024gemini15unlockingmultimodal} and GPT4-o\footnote{\url{https://openai.com/index/hello-gpt-4o}}. We share the model version details in Appendix \ref{app: modeldetails}. We use similar prompts for both datasets with minor changes in the instructions to better fit and describe each task and to resolve additional ambiguities specific to each dataset.

\subsection{Results and Discussion}
\noindent\textbf{Results for Hi-ToM.} We observe improvements in performance for all models except for the Gemini Flash model. The largest increase is seen for the GPT-4o model with an average accuracy gain of 28.13\% over the SimToM method, which slightly outperformed the chain-of-thought method. Similarly, the Llama-3-70B model saw the next largest increase of 22.5\% over the SimToM method, which slightly outperformed the baseline method. The Llama-3-8B model and the Gemini-Flash model both show slight increases over the baseline and SimToM method respectively.

Using Decompose-ToM, we observe that the performance of all models is preserved better than other methods for higher-order tasks. GPT-4o and Llama-3-70B models even see improvements of 6.6\% and 13.3\% respectively from orders 1 to 4. The Llama-3-8B and the Gemini-Flash-1.5 models show a dip in accuracy over the range. We hypothesize that larger models such as GPT-4o and Llama-3-70B can consistently perform well in the statement awareness task, whereas less capable models may fail. Conversely, the observed increase in accuracy for larger models may be attributed to the tendency of LLMs to become distracted \cite{shi2023largelanguagemodelseasily}. This distraction is likely exacerbated during lower-order reasoning due to the presence of a longer story, which introduces a greater number of potentially distracting statements. Refer to Appendix \ref{smallermodels} for a detailed discussion.

\noindent\textbf{Results for FANToM.} We observe improvements for all models over baseline methods. However, we get nearly equal performance as the SimToM method across all models but Gemini-Flash. The Decompose-ToM method when used on the Gemini-1.5-Flash model outperforms the SimToM method by 6.65\%.

The models show poorer performance on longer stories compared to short stories for both presented baselines with an average accuracy difference of 7.57\% for vanilla MC prompting, and 7.73\% for CoT prompting. The SimToM method also has an average difference of 4\% between the long and short context stories. Our method reduces the average performance gap on long and short FANToM stories to 0.9\%. These results suggest that Decompose-ToM helps language models maintain theory-of-mind-related reasoning across longer contexts while simultaneously improving performance.

\noindent\textbf{Algorithm Generalizability:} For a more detailed discussion around the generalizability of the algorithm, refer to Appendix \ref{apps:general}.

\section{Conclusion}

In this work, we introduce a novel LLM-powered inference algorithm inspired by the concepts of pretend-play and knowledge-access from cognitive psychology to enhance Theory of Mind (ToM) capabilities in LLMs. Our method decomposes complex ToM tasks into simpler sub-tasks: recursive perspective simulation and granular statement awareness, enabling models to better understand and attribute mental states in higher-order and naturalistic settings.

We evaluated our approach on the Hi-ToM and FANToM datasets, demonstrating improvements over baseline methods across both open-source and closed-source LLMs. Notably, our method showed strong performance gains in higher-order ToM tasks and effectively maintained accuracy across longer context lengths, addressing key challenges in current ToM evaluations for LLMs.

These results suggest that leveraging cognitive developmental strategies can effectively enhance the social reasoning abilities of LLMs without requiring additional model training or extensive prompt engineering. Our approach offers a flexible and generalizable framework for improving ToM performance in LLMs, which could have broad implications for the development of socially-aware AI systems. Additionally, future work can explore employing these algorithms through LLM agentic frameworks, allowing LLMs to autonomously conduct the process with a single instruction.

\section{Limitations}

A key limitation of our work is that it assumes that the agent's memory is never updated backward. For example, if a story doesn't strictly follow a chronological order by claiming midway through that an agent who hadn't been mentioned yet was present earlier (implying that they knew about what happened earlier), our method does not update this in the agent simulation. We believe that this problem of incomplete information can be resolved by disambiguating the relevant story in the pre-processing stage on the lines of our procedure for the FANToM dataset (refer to Appendix \ref{app:datasets}). However, we haven't tested this extensively and leave this for future work.

Additionally, our approach is highly computationally expensive compared to other approaches due to the need for per-statement processing, we hope to come up with more efficient methods in future work.

\section*{Acknowledgements}
The work in this paper is supported by 
by the NYUAD Center for Artificial Intelligence and Robotics, funded by Tamkeen under the NYUAD Research Institute Award CG010.

\begin{thebibliography}{38}
\providecommand{\natexlab}[1]{#1}

\bibitem[{Baron-Cohen(1995)}]{BaronCohen1995}
Simon Baron-Cohen. 1995.
\newblock \href {https://doi.org/10.7551/mitpress/4635.001.0001} {\emph{Mindblindness: An Essay on Autism and Theory of Mind}}.
\newblock The MIT Press.

\bibitem[{Baron-Cohen et~al.(1985)Baron-Cohen, Leslie, and Frith}]{Baron-Cohen-Autistic-Child-Have-ToM?}
Simon Baron-Cohen, Alan Leslie, and Uta Frith. 1985.
\newblock \href {https://doi.org/10.1016/0010-0277(85)90022-8} {Does the autistic child have a theory of mind?}
\newblock \emph{Cognition}, 21:37--46.

\bibitem[{Christensen and Michael(2012)}]{Christensen2012}
Wayne Christensen and John Michael. 2012.
\newblock \href {https://doi.org/10.1007/s11097-012-9292-9} {Ian apperly, mindreaders: the cognitive basis of theory of mind: Psychology press, 2011, 232 pages, (isbn 978-1-84169-697-3) $80.00}.
\newblock \emph{Phenomenology and the Cognitive Sciences}, 12(4):907–914.

\bibitem[{Dubey et~al.(2024)Dubey, Jauhri, Pandey, Kadian, Al-Dahle, Letman, Mathur, Schelten, Yang, Fan et~al.}]{dubey2024llama3herdmodels}
Abhimanyu Dubey, Abhinav Jauhri, Abhinav Pandey, Abhishek Kadian, Ahmad Al-Dahle, Aiesha Letman, Akhil Mathur, Alan Schelten, Amy Yang, Angela Fan, et~al. 2024.
\newblock The llama 3 herd of models.
\newblock \emph{arXiv preprint arXiv:2407.21783}.

\bibitem[{Frank and Goodman(2012)}]{frank2012predicting}
Michael~C Frank and Noah~D Goodman. 2012.
\newblock Predicting pragmatic reasoning in language games.
\newblock \emph{Science}, 336(6084):998--1002.

\bibitem[{Gandhi et~al.(2023)Gandhi, Fränken, Gerstenberg, and Goodman}]{gandhi2023understanding}
Kanishk Gandhi, Jan-Philipp Fränken, Tobias Gerstenberg, and Noah~D. Goodman. 2023.
\newblock \href {https://arxiv.org/abs/2306.15448} {Understanding social reasoning in language models with language models}.
\newblock \emph{Preprint}, arXiv:2306.15448.

\bibitem[{Jiang et~al.(2024)Jiang, Manoranjan, Salam, and Celiktutan}]{jiang2024towards}
Jian Jiang, Viswonathan Manoranjan, Hanan Salam, and Oya Celiktutan. 2024.
\newblock Towards generalised and incremental bias mitigation in personality computing.
\newblock \emph{IEEE Transactions on Affective Computing}.

\bibitem[{Kavanaugh(2006)}]{Kavanaugh2006-dk}
R~D Kavanaugh. 2006.
\newblock Pretend play and theory of mind.
\newblock In L~S Balter~C, editor, \emph{Child psychology: A handbook of contemporary issues}, pages 153--166. Psychology Press.

\bibitem[{Kim et~al.(2023)Kim, Sclar, Zhou, Bras, Kim, Choi, and Sap}]{kim-etal-2023-fantom}
Hyunwoo Kim, Melanie Sclar, Xuhui Zhou, Ronan Bras, Gunhee Kim, Yejin Choi, and Maarten Sap. 2023.
\newblock \href {https://doi.org/10.18653/v1/2023.emnlp-main.890} {{FANT}o{M}: A benchmark for stress-testing machine theory of mind in interactions}.
\newblock In \emph{Proceedings of the 2023 Conference on Empirical Methods in Natural Language Processing}, pages 14397--14413, Singapore. Association for Computational Linguistics.

\bibitem[{Kinderman et~al.(1998)Kinderman, Dunbar, and Bentall}]{Kinderman1998}
Peter Kinderman, Robin Dunbar, and Richard~P. Bentall. 1998.
\newblock \href {https://doi.org/10.1111/j.2044-8295.1998.tb02680.x} {Theory‐of‐mind deficits and causal attributions}.
\newblock \emph{British Journal of Psychology}, 89(2):191–204.

\bibitem[{Kosinski(2023)}]{kosinski2023theory}
Michal Kosinski. 2023.
\newblock \href {https://arxiv.org/abs/2302.02083} {Theory of mind may have spontaneously emerged in large language models}.
\newblock \emph{Preprint}, arXiv:2302.02083.

\bibitem[{Kwon et~al.(2023)Kwon, Li, Zhuang, Sheng, Zheng, Yu, Gonzalez, Zhang, and Stoica}]{kwon2023efficient}
Woosuk Kwon, Zhuohan Li, Siyuan Zhuang, Ying Sheng, Lianmin Zheng, Cody~Hao Yu, Joseph~E. Gonzalez, Hao Zhang, and Ion Stoica. 2023.
\newblock Efficient memory management for large language model serving with pagedattention.
\newblock In \emph{Proceedings of the ACM SIGOPS 29th Symposium on Operating Systems Principles}.

\bibitem[{Le et~al.(2019)Le, Boureau, and Nickel}]{le-etal-2019-revisiting}
Matthew Le, Y-Lan Boureau, and Maximilian Nickel. 2019.
\newblock \href {https://doi.org/10.18653/v1/D19-1598} {Revisiting the evaluation of theory of mind through question answering}.
\newblock In \emph{Proceedings of the 2019 Conference on Empirical Methods in Natural Language Processing and the 9th International Joint Conference on Natural Language Processing (EMNLP-IJCNLP)}, pages 5872--5877, Hong Kong, China. Association for Computational Linguistics.

\bibitem[{Leslie(1987)}]{Leslie1987}
Alan~M. Leslie. 1987.
\newblock \href {https://doi.org/10.1037/0033-295x.94.4.412} {Pretense and representation: The origins of “theory of mind.”}.
\newblock \emph{Psychological Review}, 94(4):412–426.

\bibitem[{Li et~al.(2023)Li, Waleed, and Salam}]{li2023survey}
Jialin Li, Alia Waleed, and Hanan Salam. 2023.
\newblock A survey on personalized affective computing in human-machine interaction.
\newblock \emph{arXiv preprint arXiv:2304.00377}.

\bibitem[{Liddle and Nettle(2006)}]{HitomSocialCompetenceChildren}
Bethany Liddle and Daniel Nettle. 2006.
\newblock \href {https://doi.org/10.1556/JCEP.4.2006.3-4.3} {Higher-order theory of mind and social competence in school-age children}.
\newblock \emph{Journal of Cultural and Evolutionary Psychology Akadémiai Kiadó}, 43:1589--5254.

\bibitem[{Lillard(1993)}]{Lillard1993-pw}
Angeline~S Lillard. 1993.
\newblock Pretend play skills and the child's theory of mind.
\newblock \emph{Child Dev.}, 64(2):348.

\bibitem[{Ma et~al.(2023)Ma, Sansom, Peng, and Chai}]{ma-etal-2023-towards-holistic}
Ziqiao Ma, Jacob Sansom, Run Peng, and Joyce Chai. 2023.
\newblock \href {https://doi.org/10.18653/v1/2023.findings-emnlp.72} {Towards a holistic landscape of situated theory of mind in large language models}.
\newblock In \emph{Findings of the Association for Computational Linguistics: EMNLP 2023}, pages 1011--1031, Singapore. Association for Computational Linguistics.

\bibitem[{Mitchell and Phillips(2015)}]{OverlappingRelationshipEmotionPerception&TOM}
{Rachel L C} Mitchell and {Louise H} Phillips. 2015.
\newblock \href {https://doi.org/10.1016/j.neuropsychologia.2015.02.018} {The overlapping relationship between emotion perception and theory of mind}.
\newblock \emph{Neuropsychologia}, 70:1--10.
\newblock Copyright {\textcopyright} 2015 Elsevier Ltd. All rights reserved.

\bibitem[{Moghaddam and Honey(2023)}]{moghaddam2023boosting}
Shima~Rahimi Moghaddam and Christopher~J. Honey. 2023.
\newblock \href {https://arxiv.org/abs/2304.11490} {Boosting theory-of-mind performance in large language models via prompting}.
\newblock \emph{Preprint}, arXiv:2304.11490.

\bibitem[{Nematzadeh et~al.(2018)Nematzadeh, Burns, Grant, Gopnik, and Griffiths}]{nematzadeh-etal-2018-evaluating}
Aida Nematzadeh, Kaylee Burns, Erin Grant, Alison Gopnik, and Tom Griffiths. 2018.
\newblock \href {https://doi.org/10.18653/v1/D18-1261} {Evaluating theory of mind in question answering}.
\newblock In \emph{Proceedings of the 2018 Conference on Empirical Methods in Natural Language Processing}, pages 2392--2400, Brussels, Belgium. Association for Computational Linguistics.

\bibitem[{OpenAI(2023)}]{openai2023gpt4}
OpenAI. 2023.
\newblock \href {https://arxiv.org/abs/2303.08774} {Gpt-4 technical report}.
\newblock \emph{arXiv preprint arXiv:2303.08774}.

\bibitem[{Premack and Woodruff(1978)}]{Premack1978}
David Premack and Guy Woodruff. 1978.
\newblock \href {https://doi.org/10.1017/s0140525x00076512} {Does the chimpanzee have a theory of mind?}
\newblock \emph{Behavioral and Brain Sciences}, 1(4):515–526.

\bibitem[{Qu et~al.(2015)Qu, Shen, Chee, and Chen}]{Qu2015}
Li~Qu, Pinxiu Shen, Yu~Yan Chee, and Luxi Chen. 2015.
\newblock \href {https://doi.org/10.1111/sode.12116} {Teachers’ theory‐of‐mind coaching and children’s executive function predict the training effect of sociodramatic play on children’s theory of mind}.
\newblock \emph{Social Development}, 24(4):716–733.

\bibitem[{Reid et~al.(2024)Reid, Savinov, Teplyashin, Lepikhin, Lillicrap, Alayrac, Soricut, Lazaridou, Firat, Schrittwieser et~al.}]{geminiteam2024gemini15unlockingmultimodal}
Machel Reid, Nikolay Savinov, Denis Teplyashin, Dmitry Lepikhin, Timothy Lillicrap, Jean-baptiste Alayrac, Radu Soricut, Angeliki Lazaridou, Orhan Firat, Julian Schrittwieser, et~al. 2024.
\newblock Gemini 1.5: Unlocking multimodal understanding across millions of tokens of context.
\newblock \emph{arXiv preprint arXiv:2403.05530}.

\bibitem[{Salam et~al.(2023)Salam, Celiktutan, Gunes, and Chetouani}]{salam2023automatic}
Hanan Salam, Oya Celiktutan, Hatice Gunes, and Mohamed Chetouani. 2023.
\newblock Automatic context-aware inference of engagement in hmi: A survey.
\newblock \emph{IEEE Transactions on Affective Computing}.

\bibitem[{Salam et~al.(2022)Salam, Manoranjan, Jiang, and Celiktutan}]{salam2022learning}
Hanan Salam, Viswonathan Manoranjan, Jian Jiang, and Oya Celiktutan. 2022.
\newblock Learning personalised models for automatic self-reported personality recognition.
\newblock In \emph{Understanding Social Behavior in Dyadic and Small Group Interactions}, pages 53--73. PMLR.

\bibitem[{Sap et~al.(2022)Sap, Le~Bras, Fried, and Choi}]{sap-etal-2022-neural}
Maarten Sap, Ronan Le~Bras, Daniel Fried, and Yejin Choi. 2022.
\newblock \href {https://doi.org/10.18653/v1/2022.emnlp-main.248} {Neural theory-of-mind? on the limits of social intelligence in large {LM}s}.
\newblock In \emph{Proceedings of the 2022 Conference on Empirical Methods in Natural Language Processing}, pages 3762--3780, Abu Dhabi, United Arab Emirates. Association for Computational Linguistics.

\bibitem[{Sclar et~al.(2023)Sclar, Kumar, West, Suhr, Choi, and Tsvetkov}]{sclar-etal-2023-minding}
Melanie Sclar, Sachin Kumar, Peter West, Alane Suhr, Yejin Choi, and Yulia Tsvetkov. 2023.
\newblock \href {https://doi.org/10.18653/v1/2023.acl-long.780} {Minding language models{'} (lack of) theory of mind: A plug-and-play multi-character belief tracker}.
\newblock In \emph{Proceedings of the 61st Annual Meeting of the Association for Computational Linguistics (Volume 1: Long Papers)}, pages 13960--13980, Toronto, Canada. Association for Computational Linguistics.

\bibitem[{Shanton and Goldman(2010)}]{Shanton2010}
Karen Shanton and Alvin Goldman. 2010.
\newblock \href {https://doi.org/10.1002/wcs.33} {Simulation theory}.
\newblock \emph{WIREs Cognitive Science}, 1(4):527–538.

\bibitem[{Shapira et~al.(2023)Shapira, Levy, Alavi, Zhou, Choi, Goldberg, Sap, and Shwartz}]{shapira2023cleverhansneuraltheory}
Natalie Shapira, Mosh Levy, Seyed~Hossein Alavi, Xuhui Zhou, Yejin Choi, Yoav Goldberg, Maarten Sap, and Vered Shwartz. 2023.
\newblock \href {https://arxiv.org/abs/2305.14763} {Clever hans or neural theory of mind? stress testing social reasoning in large language models}.
\newblock \emph{Preprint}, arXiv:2305.14763.

\bibitem[{Shi et~al.(2023)Shi, Chen, Misra, Scales, Dohan, Chi, Schärli, and Zhou}]{shi2023largelanguagemodelseasily}
Freda Shi, Xinyun Chen, Kanishka Misra, Nathan Scales, David Dohan, Ed~Chi, Nathanael Schärli, and Denny Zhou. 2023.
\newblock \href {https://arxiv.org/abs/2302.00093} {Large language models can be easily distracted by irrelevant context}.
\newblock \emph{Preprint}, arXiv:2302.00093.

\bibitem[{Ullman(2023)}]{ullman2023largelanguagemodelsfail}
Tomer Ullman. 2023.
\newblock \href {https://arxiv.org/abs/2302.08399} {Large language models fail on trivial alterations to theory-of-mind tasks}.
\newblock \emph{Preprint}, arXiv:2302.08399.

\bibitem[{Wei et~al.(2022)Wei, Wang, Schuurmans, Bosma, Ichter, Xia, Chi, Le, and Zhou}]{wei2022chain}
Jason Wei, Xuezhi Wang, Dale Schuurmans, Maarten Bosma, Brian Ichter, Fei Xia, Ed~H Chi, Quoc~V Le, and Denny Zhou. 2022.
\newblock Chain of thought prompting elicits reasoning in large language models.
\newblock \emph{Advances in Neural Information Processing Systems}, 35:24824--24837.

\bibitem[{Wellman and Liu(2004)}]{Wellman2004}
Henry~M. Wellman and David Liu. 2004.
\newblock \href {https://doi.org/10.1111/j.1467-8624.2004.00691.x} {Scaling of theory‐of‐mind tasks}.
\newblock \emph{Child Development}, 75(2):523–541.

\bibitem[{Wilf et~al.(2024)Wilf, Lee, Liang, and Morency}]{wilf-etal-2024-think}
Alex Wilf, Sihyun Lee, Paul~Pu Liang, and Louis-Philippe Morency. 2024.
\newblock \href {https://aclanthology.org/2024.acl-long.451} {Think twice: Perspective-taking improves large language models{'} theory-of-mind capabilities}.
\newblock In \emph{Proceedings of the 62nd Annual Meeting of the Association for Computational Linguistics (Volume 1: Long Papers)}, pages 8292--8308, Bangkok, Thailand. Association for Computational Linguistics.

\bibitem[{Wu et~al.(2023)Wu, He, Jia, Mihalcea, Chen, and Deng}]{wu-etal-2023-hi}
Yufan Wu, Yinghui He, Yilin Jia, Rada Mihalcea, Yulong Chen, and Naihao Deng. 2023.
\newblock \href {https://doi.org/10.18653/v1/2023.findings-emnlp.717} {Hi-{T}o{M}: A benchmark for evaluating higher-order theory of mind reasoning in large language models}.
\newblock In \emph{Findings of the Association for Computational Linguistics: EMNLP 2023}, pages 10691--10706, Singapore. Association for Computational Linguistics.

\bibitem[{Xu et~al.(2024)Xu, Zhao, Zhu, Du, and He}]{xu-etal-2024-opentom}
Hainiu Xu, Runcong Zhao, Lixing Zhu, Jinhua Du, and Yulan He. 2024.
\newblock \href {https://aclanthology.org/2024.acl-long.466} {{O}pen{T}o{M}: A comprehensive benchmark for evaluating theory-of-mind reasoning capabilities of large language models}.
\newblock In \emph{Proceedings of the 62nd Annual Meeting of the Association for Computational Linguistics (Volume 1: Long Papers)}, pages 8593--8623, Bangkok, Thailand. Association for Computational Linguistics.

\end{thebibliography}

\appendix
\newpage
\section{Generalizability}
\label{apps:general}

We've designed Decompose-ToM in a manner that makes it highly generalizable, and easy to use in a plug-and-play manner. The core Decompose-ToM function simply requires a question, options, and a story. Users can optionally pass any instructions the model should follow when interpreting the story, and the unit of an information chunk.
By information chunk, we refer to the atomic unit of information that an agent either knows or doesn't. For example, the information chunk in Hi-ToM was a sentence, and the chunk in FANToM was a dialogue. By default, the unit is a sentence.

However, although the system is generalizable the user will likely have to tune the simulation, and question-answering prompts to get optimal results. In future work, we hope to have the system be able to dynamically define information chunks, and understand the required processing steps automatedly.

\section{Algorithms}
\label{app:algos}
Algorithms \ref{alg:SimulationProcess} and \ref{alg:StoryUpdate} describe the algorithms for the simulation process and story update proposed in this work.
\begin{algorithm}
\caption{Simulation Process}
\label{alg:SimulationProcess}
\begin{algorithmic}[1]
\small
\State $Agent \leftarrow \text{GET\_AGENT(Question)}$
\While{$Agent \text{ is not Narrator}$}
    \State $W \leftarrow \text{SETUP\_WORLD(Story)}$
    \State $Agent \leftarrow \text{GET\_AGENT(Question)}$
    \State $Qn \leftarrow \text{REPHRASE\_QN(Agent, Question)}$
    \State $Story, W \leftarrow \text{SIM(Story, W, Agent)}$
\EndWhile
\State $\text{ANSWER(Story, Agent, Qn, Choices)}$
\end{algorithmic}
\end{algorithm}

\begin{algorithm}
\caption{Story Update Simulation}
\label{alg:StoryUpdate}
\begin{algorithmic}[1]
\small 
\State $CurrStory \leftarrow \text{EMPTY}$
\State $Update \leftarrow \text{EMPTY}$
\For{$P \text{ in } STORY \text{ given } CurrStory$}
    \If{$Agent \text{ aware of } P$}
        \State Add $P$ to $Update$
    \EndIf
    \State Add $P$ to $CurrStory$
    \State $W \leftarrow \text{UPDATE}(W, P)$
\EndFor
\State \Return $Update, W$
\end{algorithmic}
\end{algorithm}
\section{Disambiguating Datasets and Examples}
\label{app:datasets}
\paragraph{Hi-ToM}

 The Hi-ToM dataset consists of some unintentional ambiguities, which are also present in precursor ToM datasets such as ToMi \cite{le-etal-2019-revisiting}. These ambiguities do not allow us to ascertain the actual ToM capabilities of language models. For ToMi, these ambiguities have been identified and resolved \citep{sclar-etal-2023-minding}, however, this is not the case for Hi-ToM. For instance, a story might state, 'Bob is in the living room' and 'The beans are in the suitcase,' but never explicitly mention that the suitcase is in the living room. Since the stories are generated through fixed-structure templates, we can use regular expression parsing to form a dictionary of locations and parent locations. Ultimately, we disambiguate the locations by adding disambiguating lines at the start of the story (e.g. ``The suitcase is in the living room'').

\paragraph{FANToM }

Unlike Hi-ToM, FANToM doesn't have markers of entry into conversation at the beginning of the story and this may lead to problems for our algorithm since it does a simple forward pass over the statements. Thus, we preprocess FANToM stories to initialize the world state with the agents initially in a conversation to be in an "in-conversation" location state. We do this by simply detecting the agents in conversation at the beginning using a single LLM prompt, and setting up the initial world-state accordingly.

\begin{tcolorbox}[colback=gray!10, colframe=black, boxrule=0.5mm, sharp corners, width=\linewidth, title=HiToM Example]
Read the following story and answer the multiple-choice question. Please provide answer without explanations.

\textbf{Story:} \\
1. Emma, Hannah, Liam, Nathan and Carter entered the study.\\
2. Emma saw a cat.\\
3. The tomato is in the red\_container.\\
4. Emma made no movements and stayed in the study for 1 minute.\\
5. Emma exited the study.\\
6. Hannah made no movements and stayed in the study for 1 minute.\\
7. Hannah exited the study.\\
8. Carter dislikes the banana.\\
9. Liam moved the tomato to the green\_drawer.\\
10. Liam exited the study.\\
11. Nathan made no movements and stayed in the study for 1 minute.\\
12. Nathan exited the study.\\
13. Carter made no movements and stayed in the study for 1 minute.\\
14. Carter exited the study.\\
15. Emma, Hannah, Liam, Nathan and Carter entered the waiting\_room.\\
16. Nathan publicly claimed that the tomato is in the blue\_bottle.\\
17. Carter privately told Emma that the tomato is in the green\_drawer.\\

\textbf{Question:} \\
Where does Carter think Hannah thinks Liam thinks Emma thinks the tomato is?

\textbf{Choices:}\\
A. blue\_bathtub, B. red\_drawer, C. green\_bathtub, D. green\_envelope, E. blue\_cupboard,\\ 
F. green\_box, G. blue\_drawer, H. green\_pantry, I. green\_cupboard, J. blue\_treasure\_chest,\\ 
K. red\_bottle, L. red\_container, M. green\_bucket, N. green\_drawer, O. blue\_bottle.

\end{tcolorbox}
\begin{tcolorbox}[
    colframe=black!75!white, 
    colback=white!95!black, 
    title=FANToM Example,
    boxrule=0.8pt,
    arc=4pt,
    left=6pt,
    right=6pt,
    top=6pt,
    bottom=6pt
]
Gianna: Guys, I've really enjoyed sharing our pet stories, but I need to excuse myself. I need to change clothes for a meeting later. Talk to you later!\\
Sara: Sure thing, Gianna. Take care!\\
Javier: Catch you later, Gianna.\\
Sara: So Javier, have you ever tried training Bruno?\\
Javier: Yes, I did actually. It was a challenge at times, but rewarding nevertheless. How about you? Did you try training Snowflake?\\
Sara: Oh gosh, trying to train a cat is a whole different ball game. But I did manage to teach her a few commands and tricks. She was quite an intelligent little furball.\\
Gianna: Hey guys, I'm back, couldn't miss out on more pet stories. Speaking of teaching and training pets, it is amazing how that further strengthens the bond between us and our pets, right?\\
Sara: Absolutely, Gianna! The fact that they trust us enough to learn from us is really special.\\
Javier: I can't agree more. I believe that's one of the ways Bruno conveyed his love and trust towards me. It also gave me a sense of responsibility towards him.\\
Gianna: Just like Chirpy. Once she began to imitate me, we connected in a way I never imagined. She would repeat words that I was studying for exams and that somehow made studying less stressful.\\
Javier: Pets are indeed lifesavers in so many ways.\\
Sara: They bring so much joy and laughter too into our lives. I mean, imagine a little kitten stuck in a vase! I couldn't have asked for a better stress buster during my college days.\\
Gianna: Totally, they all are so amazing in their unique ways. It's so nice to have these memories to look back on.
\end{tcolorbox}

\section{Model Details}
\label{app: modeldetails}

We run Meta's Llama-3-70B, and Llama-3-8B models using vLLM \cite{kwon2023efficient} on 4 A100 GPUs. We use OpenAI's gpt-4o-2024-08-06 model and Gemini-1.5-Flash-001 through the publicly available APIs.

\section{Discussion on Degradation across order for Smaller Language Models}
\label{smallermodels}

We observe that LLMs such as Llama-3-8b, and Gemini-1.5-Flash suffer degradation across orders when we use our method. On evaluating this phenomenon we identify that this is caused due to the snowballing effect of errors in the knowledge-awareness task. That is, if the LLM incorrectly labels a statement x, which the agent should have known, as unknown, at simulation step n. Then, at every subsequent simulation step n+i, the LLM will have incomplete information to answer the statement to answer the statements after x. If statement X is crucial, such as information about an agent entering or leaving a room, this error will erase all statements following x at step n+1. 

Additionally, even if statement x is not crucial to the currently simulated agent, it may have important context for an agent that will be simulated subsequently. Thus, the simulation is highly sensitive to errors in the knowledge-awareness step. As the number of such steps required increases with higher orders of ToM being tested, the probability of snowballing errors increases, thus reducing the performance of the model.

\section{Implementing SimToM}
\label{simtom}

We implement SimToM by handcrafting prompts for both tasks, by taking guidance from the original SimToM prompts for the ToMi dataset, and combining it with our handcrafted prompts for the Decompose-ToM method. We conduct limited prompt-tuning and answer validation to ensure performance isn't degraded due to issues such as parsing errors.

\section{Prompts}
\label{app:prompts}
In the following, we present all of the prompts that were used in this work. 

\begin{tcolorbox}[colframe=black!75!white, colback=white!95!black, title=Agent Identification Prompt]
Based on the given question, which agent's belief or perspective do we want to find first? Use the given rules to name the agent: 

\textbf{Rules:}
\begin{itemize}
    \item If the question does not mention the name of any agents, the answer should be Narrator.
    \item Otherwise, output the primary agent's name. (Pronouns such as you/I/we/they/us aren't agent names and should not be outputted)
\end{itemize}

\textbf{Examples:}

Question: Where does Alex think Raj looks for the jam? \\
Agent Name: Alex

Question: Where do I think Sam thinks the ladder is? \\
Agent Name: Sam

Question: Where does Ava think Sophie thinks Sam thinks Brad thinks the cookie is? \\
Agent Name: Ava

Question: Where is the ladder? \\
Agent Name: Narrator

Question: Where do they think the ladder is? \\
Agent Name: Narrator

\textbf{Task:} \\
Question: \{question\} \\
Agent Name: 
\end{tcolorbox}


\begin{tcolorbox}[colframe=black!75!white, colback=white!95!black, title=Question-Reframing Prompt]
Reframe the question's perspective as if it was being asked directly to \{agent\_name\} by framing another agent as the subject of the question. Don't mention \{agent\_name\}'s name or use pronouns referring to them, instead make the question direct by removing their perspective. If there are no agents that can be made the subject, make it a direct question (Example: Where is X?) Only use 'you' when it's necessary and there are no other agents that can be framed as the subject. Output just the question and nothing else.

\textbf{Examples:}

Question: Where does \{agent\_name\} think Alex will look for the chocolate? \\
New Question: Where will Alex look for the chocolate?

Question: Where does \{agent\_name\} find the apple? \\
New Question: Where is the apple?

Question: Where does \{agent\_name\} think Brandon thinks Cody thinks the banana is? \\
New Question: Where does Brandon think Cody thinks the banana is?

\textbf{Task:} \\
Question: \{question\} \\
New Question: 
\end{tcolorbox}


\begin{tcolorbox}[colframe=black!75!white, colback=white!95!black, title=Knowledge Awareness Prompt (Hi-ToM)]
This is a given story: \{story\} \newline
The story is sequential with each statement happening after the previous one (if the statement is an event). This is the next statement in the story: Statement: \{part\}. Your task is to indicate whether \{agent\} knows about the statement happening, using the following rules:

\textbf{Rules:}
\begin{itemize}
    \item The agent \{agent\} knows of any statement that mentions their own actions.
    \item The agent \{agent\} knows of a statement if the statement happens in the same location as them.
    \item The agent \{agent\} knows of statements that indicate another agent leaving a location.
    \item The agent \{agent\} does NOT know of a statement if they have left the location where the event occurs or are not in the same location as the agent involved in the statement.
    \item The agent \{agent\} only knows of a private communication if they are involved in it.
    \item The agent \{agent\} is aware of all public communications.
    \item If a statement can be interpreted ambiguously, then say yes.
\end{itemize}
\textbf{Answer:} 
\end{tcolorbox}

\begin{tcolorbox}[colframe=black!75!white, colback=white!95!black, title=Knowledge Awareness Prompt (FANToM)]
This is a given conversation: \{story\} \newline
The story is sequential with each dialogue happening after the previous one. This is the next dialogue in the story: Dialogue: \{part\}. Your task is to indicate whether \{agent\} knows about the dialogue, using the following rules:

\textbf{Rules:}
\begin{itemize}
    \item The agent \{agent\} knows a dialogue if they are in the same location or conversation.
    \item The agent \{agent\} knows all dialogues they say themselves.
    \item If \{agent\}'s location is unclear or not provided, assume they know of the dialogue.
\end{itemize}
\textbf{Answer:} 
\end{tcolorbox}


\begin{tcolorbox}[colframe=black!75!white, colback=white!95!black, title=World Model Updation Prompt (Hi-ToM)]
This is the current world state, that holds the current world location of all the agents: \newline
World State: \{glob\_world\_model\}. Please update it relevantly (if needed) after the given statement: \{part\}.

\textbf{Follow the rules in completing the task:}
\begin{itemize}
    \item No updates are needed if an agent does not enter or exit a location in the given statement.
    \item An agent exits a location only when mentioned in the given statement. In that case, add the agent to the location "Unknown" and remove them from their original location.
    \item In case an update isn't needed return the given world state. Only update the state for agents and not objects.
    \item Ensure that no agent is in 2 locations, and only in the correct location.
\end{itemize}
Use the square brackets appropriately to indicate the agents inside a location. Only return the world state in the given format and no other text. \newline
\textbf{Answer:} World State: 
\end{tcolorbox}

\begin{tcolorbox}[colframe=black!75!white, colback=white!95!black, title=World Model Updation Prompt (FANToM)]
This is the current world state, that holds the current world location of all the agents: \newline
World State: \{glob\_world\_model\}. Please update it relevantly (if needed) after the given dialogue: \{part\}.

\textbf{Follow the rules in completing the task:}
\begin{itemize}
    \item No updates are needed if an agent does not enter or exit the conversation in the given statement.
    \item An agent exits/enters a conversation only when they mention leaving/entering themselves in the given dialogue.
    \item The agent does not exit a location themselves if they only indicate someone else may be leaving.
    \item In case an update isn't needed return the given world state.
    \item Ensure that no agent is in 2 locations, and only in the correct location.
\end{itemize}
Use the square brackets appropriately to indicate the agents inside a location. Only return the world state in the given format and no other text. \newline
\textbf{Answer:} World State: 
\end{tcolorbox}


\begin{tcolorbox}[colframe=black!75!white, colback=white!95!black, title=Final Question-Answering Prompt (Hi-ToM)]
You are \{agent\}. Here is a story: \{story\}. Answer the following question about it shortly by using the following rules to guide your reasoning. Think step by step and then give your answer after Answer: 

\textbf{Question:} \{question\}, Choices: \{choices\}, 

\textbf{Rules:}
\{note.split(":")[1]\} 
\begin{itemize}
    \item All the given statements occur in sequence, with later statements occurring after earlier statements.
    \item Public and private communications can influence beliefs of agents but don't change the results of prior events that have been observed.
    \item Note that every agent tends to lie. What an agent A tells others doesn't affect A's actual belief.
    \item An agent tends to trust an agent that exited the room later than himself.
\end{itemize}
Answer the question by thoroughly considering all statements. Choose one of the choices from the given options to return your answer. Return the associated letter label of your choice (from A,B,C,D,E,F,G,H,I,J,K,L,M,N,O) alongside your choice.
\end{tcolorbox}

\begin{tcolorbox}[colframe=black!75!white, colback=white!95!black, title=Final Question-Answering Prompt (FANToM)]
You are \{agent\}. Here is a conversation between individuals who have just met from the perspective of the given agents: 

\{agents\} \newline
\{story\} \newline
Answer the following question about it shortly by using the given rules to guide your reasoning.

\textbf{Question:} \{question\}, Choices: \{choices\}, 

\textbf{Rules:}
\begin{itemize}
    \item You don't know dialogues said before you enter a room, or after you exit a room (but you may re-enter and become aware again).
    \item You don't know the answer to the question if you don't see a reference to it in the story you know.
\end{itemize}
Choose one of the choices from the given options to return your answer. Return the associated letter label of your choice (from A,B) alongside your choice.
\end{tcolorbox}

\end{document}